\documentclass[sigconf]{acmart}
\AtBeginDocument{%
  }
   
\usepackage{multirow}    
\usepackage{graphicx}    
\usepackage{colortbl}    
\usepackage{booktabs}    
\usepackage{enumitem}

\copyrightyear{2025}
\acmYear{2025}
\setcopyright{acmlicensed}
\acmConference[MM '25]{Proceedings of the 33rd ACM International Conference on Multimedia}{October 27--31, 2025}{Dublin, Ireland}
\acmBooktitle{Proceedings of the 33rd ACM International Conference on Multimedia (MM '25), October 27--31, 2025, Dublin, Ireland}
\acmDOI{10.1145/3746027.3754562}
\acmISBN{979-8-4007-2035-2/2025/10}

\begin{document}

\title[Pathology-Aware Prototype Evolution for DR Diagnosis]{Pathology-Aware Prototype Evolution via LLM-Driven Semantic Disambiguation for Multicenter Diabetic Retinopathy Diagnosis}



\author{Chunzheng Zhu}
\affiliation{%
 \institution{Hunan University}
 \city{Changsha}
 \state{Hunan}
 \country{China}}
\email{zhuchzh@hnu.edu.cn}

\author{Yangfang Lin}
\affiliation{%
 \institution{Hunan University}
 \city{Changsha}
 \state{Hunan}
 \country{China}}
\email{lyfang123@hnu.edu.cn}

\author{Jialin Shao}
\affiliation{%
 \institution{Hunan University}
 \city{Changsha}
 \state{Hunan}
 \country{China}}
\email{sjlljs176@gmail.com}

\author{Jianxin Lin}
\authornote{Corresponding author.}
\affiliation{%
 \institution{Hunan University}
 \city{Changsha}
 \state{Hunan}
 \country{China}}
\email{linjianxin@hnu.edu.cn}

\author{Yijun Wang}
\affiliation{%
 \institution{Hunan University}
 \city{Changsha}
 \state{Hunan}
 \country{China}}
\email{wyjun@hnu.edu.cn}






\renewcommand{\shortauthors}{Chunzheng Zhu, Yangfang Lin, Jialin Shao, Jianxin Lin, \& Yijun Wang}
\newcommand{\name}{HAPM\xspace}

\begin{abstract}

Diabetic retinopathy (DR) grading plays a critical role in early clinical intervention and vision preservation. Recent explorations predominantly focus on visual lesion feature extraction through data processing and domain decoupling strategies. However, they generally overlook domain-invariant pathological patterns and underutilize the rich contextual knowledge of foundation models, relying solely on visual information, which is insufficient for distinguishing subtle pathological variations. Therefore, we propose integrating fine-grained pathological descriptions to complement prototypes with additional context, thereby resolving ambiguities in borderline cases. Specifically, we propose a Hierarchical Anchor Prototype Modulation (HAPM) framework to facilitate DR grading. First, we introduce a variance spectrum-driven anchor prototype library that preserves domain-invariant pathological patterns. We further employ a hierarchical differential prompt gating mechanism, dynamically selecting discriminative semantic prompts from both LVLM and LLM sources to address semantic confusion between adjacent DR grades. Finally, we utilize a two-stage prototype modulation strategy that progressively integrates clinical knowledge into visual prototypes through a Pathological Semantic Injector (PSI) and a Discriminative Prototype Enhancer (DPE). Extensive experiments across eight public datasets demonstrate that our approach achieves pathology-guided prototype evolution while outperforming state-of-the-art methods. The code is available at \url{https://github.com/zhcz328/HAPM}.
\end{abstract}







\maketitle

\section{Introduction}
\begin{figure}[t]
    \includegraphics[width=\linewidth]{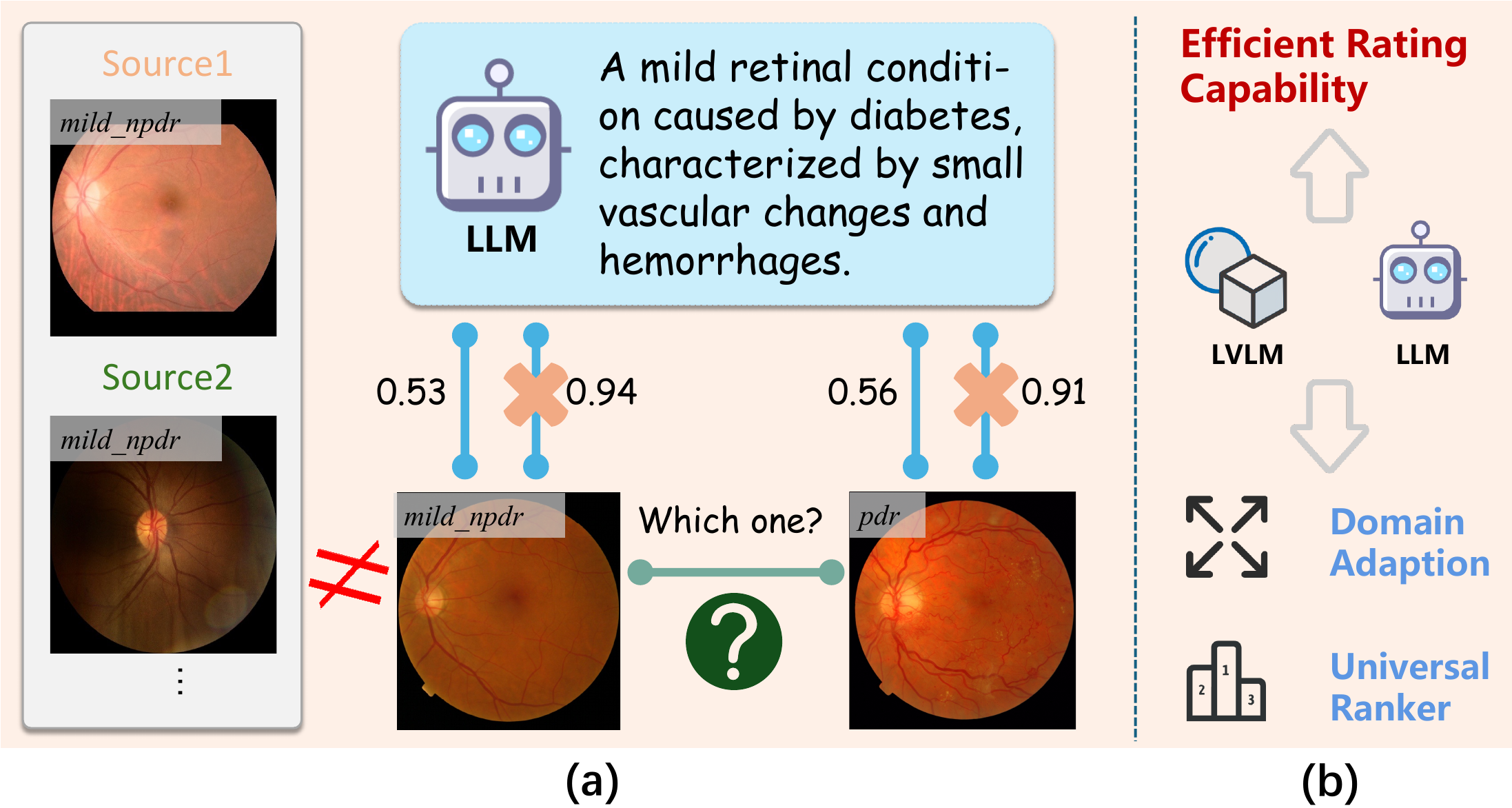}
    \caption{(a) Same DR grade appears differently across domains, and subtle differences between adjacent grades easily cause confusion. (b) Our framework combines LLM and LVLM technologies for accurate and efficient grading.}
    \label{fig:introduction}
    \vspace{-1mm}
\end{figure}

Disease grading evaluates pathological severity in medical images, guiding clinical decisions and treatment plans. In diabetic retinopathy (DR), disease progression is classified into five categories (No DR, Mild NPDR, Moderate NPDR, Severe NPDR, and PDR) according to international standards (\textit{e.g.} DRSS), requiring quantitative biomarker changes such as microaneurysm count and exudate volume for determination \cite{gulshan2016development,abramoff2016improved}. In practice, DR grading faces unique challenges: severity levels exhibit inherent semantic ambiguity, stemming from the continuity of disease progression and cross-domain heterogeneity, as shown in Figure \ref{fig:introduction}. On one hand, adjacent levels may differ only by minor morphological changes; on the other hand, retinal images of the same severity level may have significantly different texture feature distributions due to equipment differences or imaging protocol variations from various institutions, making cross-domain grading tasks more complex \cite{ghazal2020accurate,kumar2021diabetic}.


DR grading methods have witnessed significant advancements in recent years \cite{dai2021deep,akhtar2024diabetic,bilal2024nimeq}. However, existing approaches predominantly rely on data augmentation, domain decoupling or visual feature comparisons to mitigate distribution shifts. These methods fail to effectively mine the grade-invariant pathological patterns that persist across domains. In real-world applications, the following challenges arise: \textbf{1) Cross-domain sensitivity and long-tail distribution}: Imaging differences across medical centers/devices and the long-tail nature of data distribution make model localization of key lesions (\textit{e.g.} microaneurysms) susceptible to style interference \cite{li2017deeper,wang2019learning}. \textbf{2) Progression boundary ambiguity}: The high similarity between levels makes it difficult for traditional networks to distinguish minor but clinically significant pathological changes.  \textbf{3) Underutilization of foundation models}: Current approaches fail to leverage the rich contextual knowledge embedded in foundation models, overlooking the potential of pre-trained architectures, LVLMs, and LLMs to provide valuable pathological context for enhanced diagnostic accuracy. \textbf{4) Limited multimodal differentiation}: Visual features alone often prove insufficient for distinguishing subtle pathological variations, whereas integrating fine-grained textual descriptions could provide complementary context to resolve ambiguities in borderline cases.

To overcome these limitations, our preliminary investigations demonstrate that using frozen self-supervised pre-trained models to drive prototype classification, when applied to cross-domain DR datasets, results in particularly poor discrimination between adjacent severity levels. This suggests fundamental representational inadequacies in capturing the subtle pathological variations critical for accurate DR staging. While semantics can serve as an additional supervisory signal to guide prototype evolution \cite{zhang2024simple,liu2025guiding}, we observed significant overlap and intersection of prompt embeddings across different grades, causing multi-level semantic confusion between adjacent DR severity levels. Therefore, we propose a Hierarchical Anchor Prototype Modulation (HAPM) framework for DR grading through principled representational refinement.

Specifically, we first construct a variance spectrum-driven anchor prototype library by selecting representative samples from each severity class that minimize intra-class feature embedding variance, thereby establishing preliminary domain-invariant pathological prototypes. To address division ambiguity, we design a hybrid prompt architecture that bridges global case priors from vision-language models (LVLM) and large language models (LLM) with lesion-specific features. This prompt generation system combines class-level LVLM prompts with fine-grained pathological descriptions from LLMs, creating a comprehensive prompt library that captures the semantic differences between adjacent DR grades. Furthermore, we introduce a differentiated grade description mechanism that precisely captures pathological feature differences between DR grades using a template for LLM. This generates discriminative description pairs that help differentiate between easily confused categories, particularly adjacent severity levels.

Finally, we implement a two-stage prototype modulation process through the Pathological Semantic Injector (PSI) and Discriminative Prototype Enhancer (DPE), which progressively integrate diverse description features and differentiated description features into the visual prototypes. The PSI module uses an attention-based mechanism to integrate diversified description features into initial prototypes, enabling precise mapping from macro-semantic descriptions to micro-pathological regions. The DPE module then further enhances these prototypes by incorporating differentiated descriptions through an adaptive weighting mechanism that establishes clearer decision boundaries between adjacent DR severity grades. By using a frozen self-supervised pre-trained model as the backbone and designing lightweight parameter modulation modules, our approach achieves superior cross-domain performance while preserving pre-trained pathological knowledge. The main contributions of this paper can be summarized as follows:
\begin{itemize}[leftmargin=1cm]
\item We propose a variance spectrum-driven anchor prototype library that preserves domain-invariant pathological patterns through intra-class variance minimization.

\item   We develop a hierarchical differential prompt gating mechanism that dynamically selectively gates LLM-generated pathological descriptions to resolve multi-level semantic confusion, particularly in delineating adjacent grades.

\item Through our two-stage prototype modulation with the Pathological Semantic Injector and Discriminative Prototype Enhancer, we utilize pathological knowledge to refine prototypes for learning inter-class subtle differences.

\item  To our knowledge, this is the first framework that systematically integrates the DR-specific pre-trained model with multimodal foundation models to capture clinically relevant retinal disease nuances, achieving state-of-the-art performances across eight public datasets.
\end{itemize}

\begin{figure*}
    \includegraphics[width=1\linewidth]{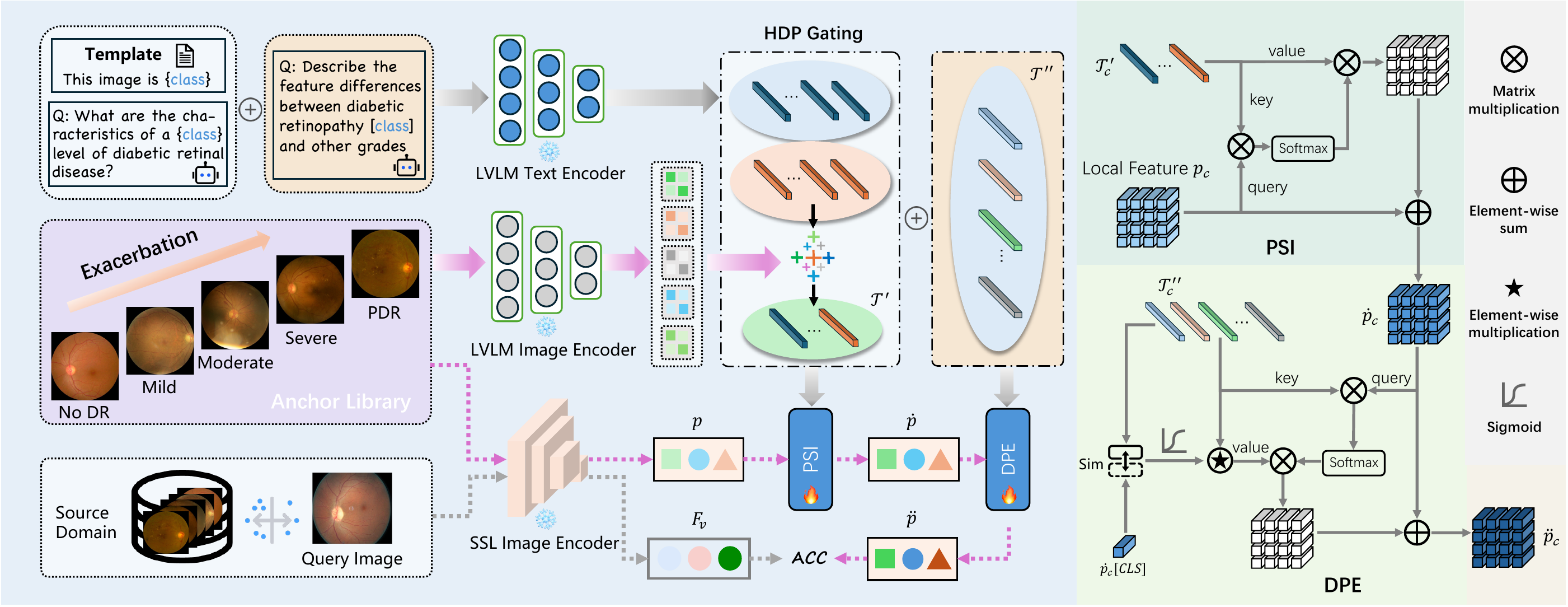}
    \caption{Overview of our method. We first build an anchor prototype library using variance spectrum analysis, then apply a Hierarchical Dynamic Prompt (HDP) Gating to select discriminative prompts. The prototypes are enhanced via two-stage modulation with the Pathological Semantic Injector (PSI) and Discriminative Prototype Enhancer (DPE) for DR grading.}
    \label{overall}
\end{figure*}

\section{Related Work}
\subsection{Diabetic Retinopathy Grading}
Recent deep learning advances have significantly improved diabetic retinopathy (DR) grading accuracy \cite{dai2021deep, kolla2021efficient}. Early CNN-based models \cite{gulshan2016development, gargeya2017automated, abramoff2016improved} extracted lesion features but couldn't model inter-organ relationships. Later research incorporated attention mechanisms to better recognize key lesion areas like microvascular abnormalities and hard exudates \cite{zhou2018multi, jiang2017automatic, kwasigroch2018deep}. Multi-stage fusion networks have brought breakthroughs in DR grading. GREEN \cite{li2019green} integrated multi-scale feature extractors. CABNet \cite{zhou2021cab} incorporated contextual information into feature learning. MIL-ViT \cite{bi2023mil} captured local pathological features using multi-instance learning. DRGen \cite{atwany2022drgen} enhanced small sample class representations through generative adversarial networks. GDRNet \cite{che2023towards} improved boundary case identification by integrating global-local relationships. The CLIP-based model CLIP-DR \cite{yu2024clip} has demonstrated strong potential in leveraging pre-trained visual-language representations for effective grading. However, these approaches neither fully harness the prior knowledge in pre-trained models nor exploit pathology-driven semantics for more discriminative grading.


\subsection{Domain Generalization in Medical Imaging}
Domain generalization (DG) techniques aim to address the issue of domain shift in medical image analysis \cite{li2017deeper, wang2019learning, zhou2020deep}. Mainstream methods include domain randomization techniques such as Mixup \cite{zhang2018mixup}, which creates synthetic training data through sample interpolation, and MixStyle \cite{zhou2021domain}, which mixes style information from different samples at the feature level; DDAIG \cite{zhou2020domain} enhances domain diversity through adversarial generation; Test-time adaptation techniques like TS \cite{sun2020test} optimize model performance adaptively during inference; Fishr \cite{rame2022fishr} innovatively uses gradient covariance regularization to alleviate domain shift by promoting gradient alignment across different domains; MDLT \cite{yang2022multi} explores multi-level domain information to improve model generalization. Recent research has also made significant progress in single-source domain generalization (SSDG) \cite{qureshi2023improving, wang2023style, zhou2022generalizable}. Chen et al. \cite{chen2022domain} proposed an enhanced framework based on contextual training; Li et al. \cite{li2023pyramid} and Liu et al. \cite{liu2021feddg} developed domain-invariant feature extraction methods based on adversarial learning. These methods often require full fine-tuning, which may disrupt pre-trained anatomical priors.


\subsection{Multimodal Prototype Learning}
Prototype learning has been widely applied in image analysis in recent years \cite{snell2017prototypical, wang2022pcl, wang2022prototype}. Recent research shows that multimodal representation learning can effectively bridge the gap between medical vision and semantics \cite{moor2023med, he2024pefomed, han2023multimodal}. The vision transformer architecture \cite{dosovitskiy2021image} provides powerful feature extraction capabilities for prototype learning. Traditional prototype networks such as ProtoNet \cite{snell2017prototypical} build class prototypes by clustering sample features, but they do not consider cross-modal knowledge transfer. New advancements in prototype learning include SemFew \cite{zhang2024simple}, which automatically aligns visual prototypes through visual-semantic evolution; and LGPN \cite{liu2025guiding}, which uses label semantics to guide prototype network learning to achieve more discriminative representations. In contrast, our framework enriches DR visual prototypes through integration of LVLM and LLM knowledge, achieving superior accuracy across domains while preserving pre-trained pathological knowledge.


\section{Problem Formulation}
\textbf{Disease Grading.}
Disease grading assesses the severity of medical conditions by analyzing pathological regions in images. This paper focuses on diabetic retinopathy (DR) grading with two primary objectives: \text{(1)} enabling the evolution of prototypes that preserve domain-invariant pathological knowledge, and \text{(2)} classifying disease severity into predefined levels $\mathcal{C}$.

\vspace{0.5mm}\noindent\textbf{Definition 1 (Prototype-based DR Grading).}
Given a fundus image $X \in \mathbb{R}^{H \times W \times 3}$, we obtain features $F_v = f_v(X)$ using a frozen self-supervised learning (SSL) visual encoder $f_v(\cdot)$. Following the DRSS criterion, we define disease severity levels as $\mathcal{C}=\{0, 1, \dots, 4\}$. Traditional prototype-based classification methods perform grading by computing the similarity between the feature $F_v$ and static prototypes $\mathcal{P}_c = \frac{1}{N_c} \sum_{i=1}^{N_c} f_v(X_i^c)$, where $X_i^c$ represents a source domain sample of grade $c \in \mathcal{C}$. Our approach achieves dynamic DR prototypes evolution process via: $\mathcal{P} \xrightarrow{\text{PSI}} \dot{\mathcal{P}} \xrightarrow{\text{DPE}} \ddot{\mathcal{P}}$
where PSI denotes Pathology Semantic Injector, DPE represents Discriminative Prototype Enhancer, and $\ddot{\mathcal{P}} = \{\ddot{\mathcal{P}}_c\}_{c \in \mathcal{C}}$ constitutes the second-level enhanced prototype set. The final  grading is determined by $\arg\max_{c \in \mathcal{C}} \textit{sim}(F_v, \ddot{\mathcal{P}}_c)$, where $\textit{sim}(\cdot,\cdot)$ is cosine similarity.

\section{Methods}

\subsection{Overview}

Figure \ref{overall} illustrates our HAPM framework. Through progressive prototype space optimization, we achieve a lightweight adjustment while maintaining cross-domain robustness and capturing fine-grained hierarchical relationships between DR progression stages. Our framework operates via three collaborative mechanisms: (1) a variance spectrum-driven anchor library preserving domain-invariant pathological patterns by selecting anchors with minimal intra-class feature variance; (2) a hierarchical dynamic prompt gating that bridges global case priors from LVLM with lesion-specific features, adaptively selecting discriminative descriptors to address confusion between adjacent DR grades; and (3) a two-stage prototype modulation that enhances visual prototypes through PSI and DPE modules, improving inter-class discriminability while maintaining intra-class consistency. This approach constructs a representation space with a hierarchical semantic structure suitable for the modulation of the progressive DR characteristics.
\begin{figure}[t]
    \includegraphics[width=0.95\linewidth]{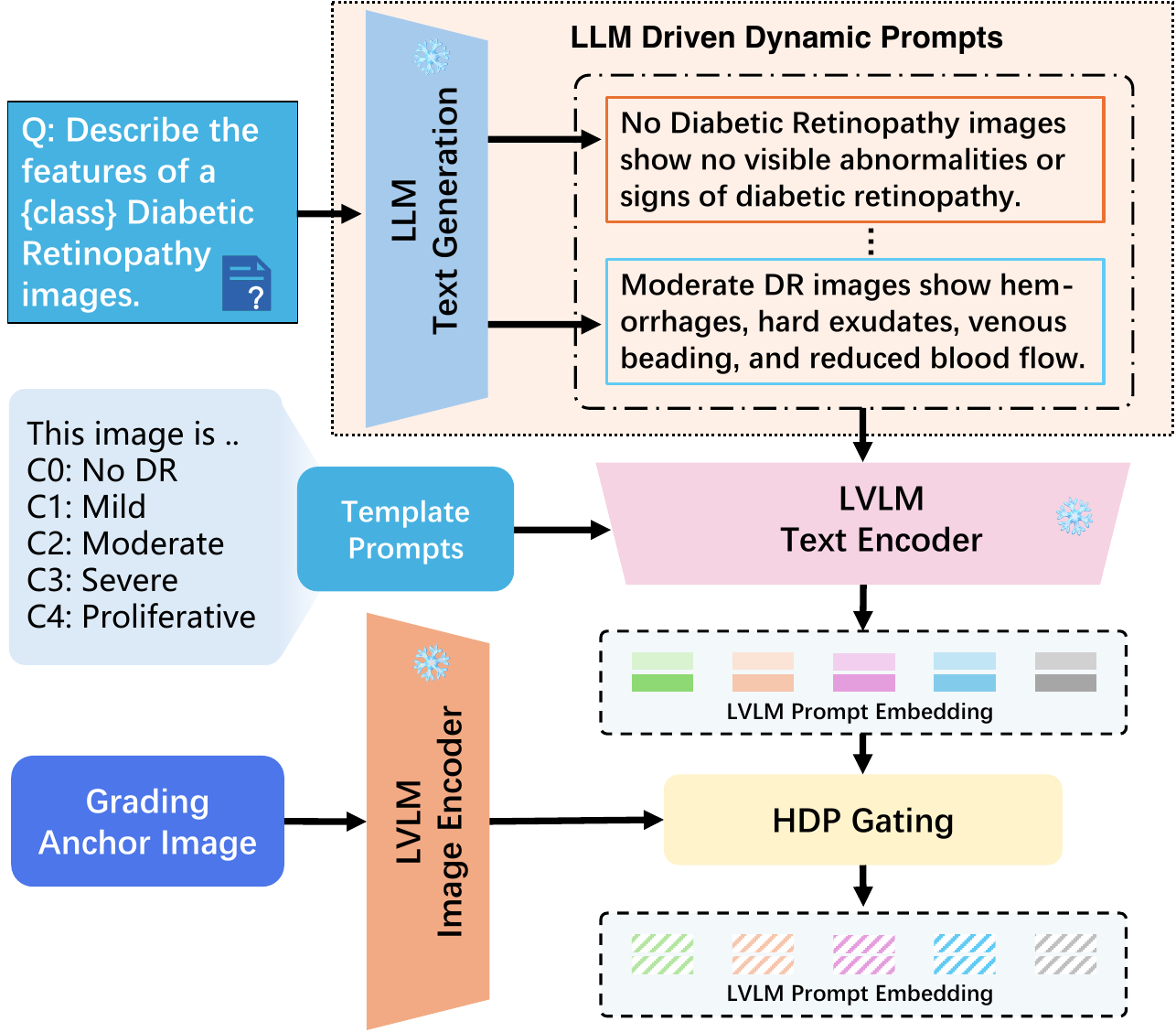}
    \caption{HDP Gating selectively filters the most discriminative prompts from both LLM and LVLM sources to reduce semantic confusion between adjacent DR grades.}
    \label{HDC}
\end{figure}

\subsection{Variance Spectrum-driven Anchor Library}
For each severity class $c \in \mathcal{C}$, we select $\alpha$ anchor images from the Eyepacs \cite{gulshan2016development} dataset as class representatives through variance spectrum analysis. The selection criterion minimizes intra-class feature embedding variance, thereby isolating high-confidence features and establishing domain-invariant pathological prototypes. Formally, using an SSL pre-trained encoder, we compute feature embeddings for all images and optimize the selection via:
\begin{equation}
A_c = \arg \min_{A \subset X_c, |A| = \alpha} \sum_{x \in A} \sigma(x)^2,
\end{equation}
where \( A_c \) represents the \( \alpha \) anchor images selected for severity level \( c \), \( X_c \) represents the set of samples for grade \( c \), and \( \sigma(x)^2 \) quantifies the feature embedding variance of image \( x \) within its class. This variance metric provides a quantitative measure of representational consistency that remains robust across different imaging devices, acquisition protocols, and institutional variations. Minimizing it ensures that the selected anchors exhibit the most stable feature representations, which is an essential property for effective cross-domain grading. These domain-invariant anchors enable robust initialization for the subsequent two-stage modulation pipeline. Finally, for all severity levels \( c \in \mathcal{C} \), the set of anchor images \( A \) for all categories can be represented as $A = \bigcup_{c \in \mathcal{C}} A_c$. This selection process identifies prototypical examples with consistent pathological manifestations across imaging conditions.

\subsection{Hierarchical Dynamic Prompt Gating}
Accurate DR grading depends on subtle semantic differences in textual prompts. Existing methods struggle with semantic variability and borderline cases. Figure \ref{HDC} shows the hybrid dynamic differential prompt (HDP) Gating framework, which adaptively selects discriminative semantic features to improve grading precision.

\vspace{0.5mm}\noindent\textbf{Hybrid Prompt Generation.} LLMs act as knowledge repositories, enabling effective processing of natural language queries \cite{chen2023see}. This paper builds on prompt learning trends \cite{Zhou_2022_CVPR,Khattak_2023_CVPR} to generate DR grade comparison prompts, addressing domain-specific language ambiguities that hinder detectors based on unified templates. We design a unified template $\mathcal{M}$ to generate DR prompts, activating CLIP-DR \cite{yu2024clip} model knowledge, unlike methods with manually defined descriptors or complex multimodal prompts \cite{deng2023anovl,yu2024clip}.

For the LLM prompt template, we formalize it as: 
$\mathcal{M}_{LLM} = $ "\texttt{Describe the typical distribution of lesions in a \{class\} diabetic retinopathy image showing}", which includes severity and optional specific detail sections. Meanwhile, we maintain the concise class description form for LVLM: $\mathcal{T}_c^{cls}=$ "\texttt{This image is \{class\}}". Based on this, we generate a five-level DR contrastive prompt library $\mathcal{T} = \cup_{c=0}^{4} \mathcal{T}_c$, where:
\begin{equation}
\mathcal{T}_c = \{t_c^{\textit{cls}}\} \cup \{t_{c,1}^{\textit{desc}}, t_{c,2}^{\textit{desc}}, \ldots, t_{c,n_c}^{\textit{desc}}\}, \quad c \in \mathcal{C}.
\end{equation}

Here, $\mathcal{T}_c^{cls} = \{t_c^{\textit{cls}}\}$ represents the class-level LVLM prompt set, and $\mathcal{T}_c^{desc} = \{t_{c,1}^{\textit{desc}}, t_{c,2}^{\textit{desc}}, \ldots, t_{c,n_c}^{\textit{desc}}\}$ represents the fine-grained pathological description set generated by LLM.

\begin{figure}
    \includegraphics[width=0.48\textwidth]{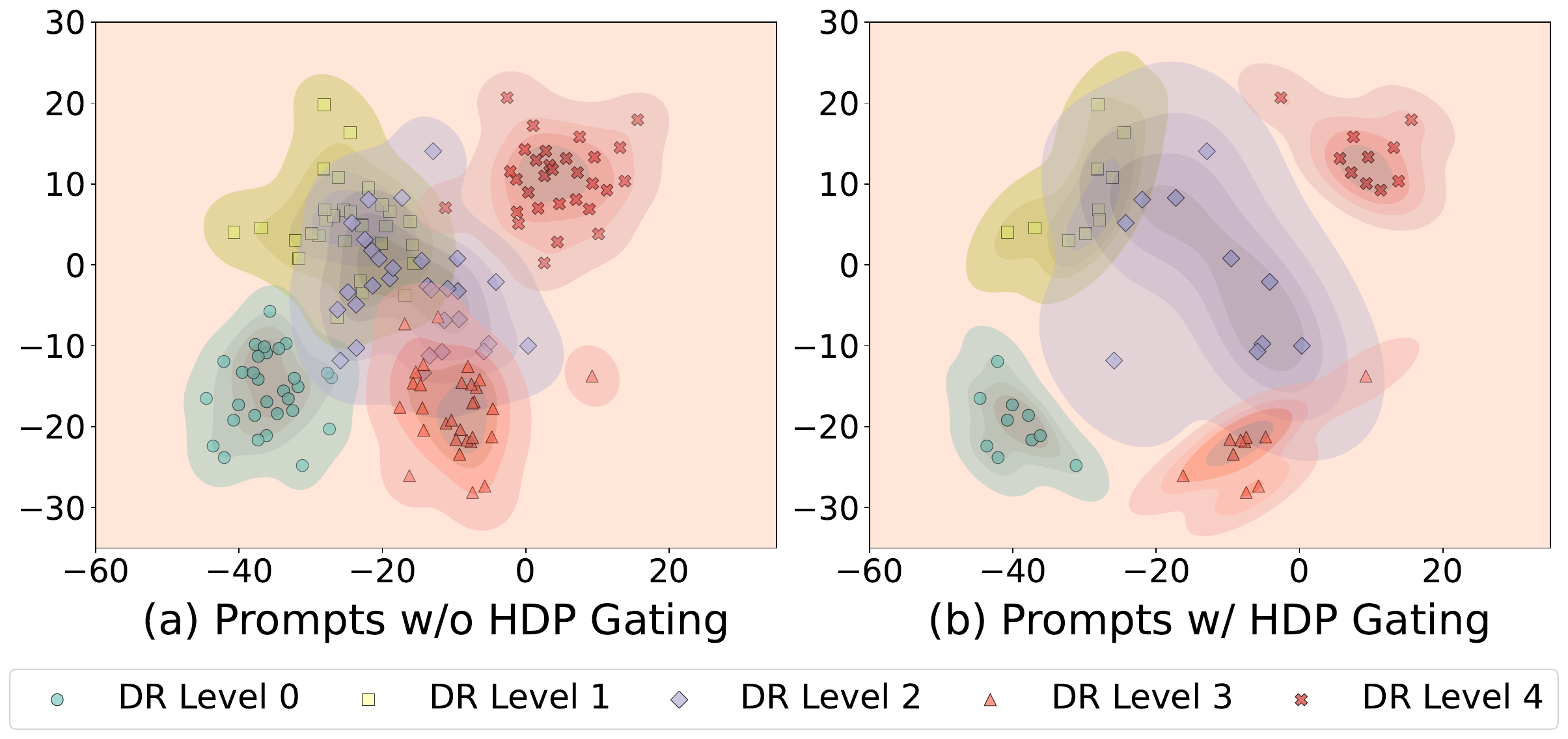}
    \caption{Selected prompts with larger inter-class distance.}
    \label{HDC Gating}
\end{figure}

\vspace{0.5mm}\noindent\textbf{Graded Semantic Confusion.}
Ideally, LVLM in the DR domain should accurately distinguish the correlation between different DR grade images $I_c$ and their corresponding level prompts $\mathcal{T}_c$, such that the similarity function $\phi(I_c, \mathcal{T}_c) \gg \phi(I_c, \mathcal{T}_{c'})$, where $c' \neq c$. However, by visualizing the semantic spaces generated by LVLM and LLM, we observed significant overlap and intersection of prompt embeddings $\mathbf{E}_{\mathcal{T}_c^{cls}}$ and $\mathbf{E}_{\mathcal{T}_c^{desc}}$ across different grades, as shown in Figure \ref{HDC Gating}. This can be formally expressed as:
\begin{equation}
\exists I_c, t_{c'} \in \mathcal{T}_{c'} \text{ s.t. } \phi(I_c, t_{c'}) > \phi(I_c, t_c), \text{ where } c' \neq c, t_c \in \mathcal{T}_c.
\end{equation}
We define this phenomenon as \textit{multi-level semantic confusion}, quantified by the confusion degree $\Delta(\mathcal{T}_c^{cls\cup desc}, \mathcal{T}_{c'}^{cls\cup desc})$, which characterizes the extent to which semantic confusion between levels $c$ and $c'$ exceeds a preset threshold. Analysis indicates that this phenomenon stems from the inherent complexity of DR pathological descriptions and the diverse semantic associations between text and images. Experiments observed significant patterns:
\begin{align}
\Delta(\mathcal{T}_1^{cls\cup desc}, \mathcal{T}_2^{cls\cup desc}) & > \Delta(\mathcal{T}_1^{cls\cup desc}, \mathcal{T}_4^{cls\cup desc}). \\
\Delta(\mathcal{T}_2^{cls\cup desc}, \mathcal{T}_3^{cls\cup desc}) & > \Delta(\mathcal{T}_2^{cls\cup desc}, \mathcal{T}_0^{cls\cup desc}).
\end{align}
This indicates that the degree of semantic confusion between adjacent DR levels (such as mild and moderate) is significantly higher than confusion between levels separated by multiple grades.

\vspace{0.5mm}\noindent\textbf{Dynamic Prompt Gating.}
To effectively address the challenge of semantic confusion in DR multi-level classification, we propose an LVLM-based discriminative prompt contextual scoring mechanism, fully leveraging LVLM's inherent discriminative ability in cross-modal representation spaces. As illustrated in Figure \ref{HDC Gating}, our approach significantly increases the inter-class distance between prompts of different DR levels, creating more distinct semantic clusters. Specifically, we introduce a dynamic level discriminative prompt gating mechanism $\mathcal{F}: \mathcal{T} \times \mathbb{R}^d \to \mathcal{T}'$, which adaptively filters the most discriminative prompt subset $\mathcal{T}' \subset \mathcal{T}$ for each anchor image $I_c$, where $|\mathcal{T}'| = N_{\text{div}}$.
Given a anchor image $I \in \mathbb{R}^{H \times W \times 3}$ and the DR grading prompt set $\mathcal{T}$, we extract corresponding features through LVLM's dual-stream encoder $F = (E_{\text{img}}, E_{\text{text}})$:
\begin{align}
\mathbf{v} &= E_{\text{img}}(I) \in \mathbb{R}^{d}, \mathbf{w}_t = E_{\text{text}}(t) \in \mathbb{R}^{d}, \quad \forall t \in \mathcal{T}.
\end{align}
In LVLM's shared semantic space, we use cosine similarity $\phi(I, t) = \frac{\langle \mathbf{v}, \mathbf{w}_t \rangle}{||\mathbf{v}||_2 \cdot ||\mathbf{w}_t||_2}$ to quantify the image-text association strength. Based on the contrastive learning principle of LVLM, we design a discriminative scoring function $\mathcal{S}_d: \mathcal{T} \times \mathbb{R}^d \to [0,1]$:
\begin{equation}
\mathcal{S}_d(t, I) = \sigma\left((\max_{c \in \mathcal{C}}\phi(I, t_c) - \frac{1}{|\mathcal{C}|-1}\sum_{c' \neq c}\phi(I, t_{c'})\right),
\end{equation}
where $\sigma(x) = \frac{1}{1+e^{-x}}$ is the Sigmoid function and $t_c \in \mathcal{T}$. A lower overall score (close to 0) indicates weaker discriminative ability of the prompt. This function enables effective distinction between closely related DR grades by prioritizing prompts that maximize similarity to the target class while minimizing similarity to non-target classes, consistent with LVLM's contrastive learning objective. Subsequently, HDP Gating selects the prompt subset $\mathcal{T}'$ with the strongest DR discriminative power:
\begin{equation}
\mathcal{T}' = \text{Top-}N_{div}(\{t \in \mathcal{T} | \mathcal{S}_d(t, I)\}).
\end{equation}
Experiments show that an appropriate $N_{div}$ value can balance information adequacy and semantic confusion risk. This mechanism deeply integrates with LVLM's vision-language alignment characteristics, precisely filtering out prompts that best represent the pathological features of specific DR levels, significantly enhancing the model's performance in recognizing clinical boundary cases.

\vspace{-0.5mm}
\subsection{Two-Stage Prototype Modulation}
Although the basic prompt $\mathcal{T'}$ contains fundamental semantic information about DR grades, it lacks sufficient class discriminability, especially in distinguishing subtle pathological differences between adjacent DR grades. Therefore, we further introduce a differentiated grade description mechanism to construct more discriminative semantic representations.

\vspace{0.5mm}\noindent\textbf{Differentiated Grade Description}. Inspired by \cite{esfandiarpoor2023follow}, which uses differentiated attribute descriptions to distinguish easily confused categories, we designed a DR-specific differentiation description template to precisely capture pathological feature differences between different DR grades:
$\mathcal{M}_{diff}$ = \texttt{"Describe the signific-\\ant pathological feature differences between diabetic retinopathy \{class1\} and \{class2\}"}.

By replacing "\texttt{\{class1\}}" and "\texttt{\{class2\}}" with any pair of DR grade names, we guide the LLM to generate multiple differentiated description sentence pairs as the differentiated description $D_{c_2}^{c_1}$ of grade $c_1$ relative to grade $c_2$. For each grade $c_n$, we combine its differentiated descriptions with all other grades to form the differentiated description set $\mathcal{T}''$ = $\{ \mathcal{T}''_{c_n} \}_{n=0}^{|\mathcal{C}|-1}$, where $\mathcal{T}''_{c_n} = \{ D_{c_i}^{c_n} \}_{c_i \in \mathcal{C} \setminus \{c_n\}}$ contains $\mathcal{|C|}*N_{\text{diff}} $ descriptions, and $N_{\text{diff}}=(|\mathcal{C}|-1)$.

\vspace{0.5mm}\noindent\textbf{Semantically Enhanced Prototype Generation.} Given the basic prompt $\mathcal{T}'$, differentiated descriptions $\mathcal{T}''$, and the original visual prototype set $\mathcal{P} = \{\mathcal{P}_c\}_{c \in \mathcal{C}}$ obtained through self-supervised learning, where each $\mathcal{P}_c \in \mathbb{R}^{N_s \times d_v}$ represents the visual prototype of a specific DR grade $c$ and $\mathcal{P} \in \mathbb{R}^{\mathcal{C} \times N_s \times d_v}$, where $N_s$ represents token length, $d_v$ represents visual feature dimension. Next, we extract two types of text features based on $\mathcal{T}'$ and $\mathcal{T}''$ through the frozen LVLM text encoder: pathological description features $\mathbf{E}' = \{\mathbf{E}'_c\}_{c \in \mathcal{C}}$, where each $\mathbf{E}'_c \in \mathbb{R}^{N_{\text{div}} \times d_t}$ represents the basic pathological description features of a specific DR grade $c$; and differentiated description features $\mathbf{E}'' = \{\mathbf{E}''_c\}_{c \in \mathcal{C}}$, where each $\mathbf{E}''_c \in \mathbb{R}^{N_{\text{diff}} \times d_t}$ represents the differentiated description features of a specific DR grade $c$ relative to other grades, where $d_t$ represents text feature dimension. 

To incorporate these rich textual semantics into visual prototypes, we design a two-stage modulation process: first, we integrate basic pathological features into prototypes through the Pathological Semantic Injector, and then further introduce differentiated features using the Discriminative Prototype Enhancer, ultimately generating refined prototypes with richer semantics and stronger discriminative ability.

\begin{table*}[t]
\centering
\renewcommand{\arraystretch}{0.4}
\setlength{\tabcolsep}{4pt}
\footnotesize
\caption{ESDG performance comparison of different methods for DR grading across multiple datasets. The best performance is highlighted in \textbf{\cellcolor{red!30}bold red}, with the second and third best highlighted in \textbf{\cellcolor{green!30}green} and \textbf{\cellcolor{yellow!30}yellow}, respectively.}
\label{tab:results}
\resizebox{0.94\textwidth}{!}{
\begin{tabular}{p{2.5cm}||cc||cc||cc||cc||cc||cc||cc}
\toprule[1.5pt]
\multirow{2}{*}{\footnotesize\sffamily\bfseries Method} & \multicolumn{2}{c||}{\footnotesize\sffamily\bfseries APTOS} & \multicolumn{2}{c||}{\footnotesize\sffamily\bfseries DeepDR} & \multicolumn{2}{c||}{\footnotesize\sffamily\bfseries FGADR} & \multicolumn{2}{c||}{\footnotesize\sffamily\bfseries IDRID} & \multicolumn{2}{c||}{\footnotesize\sffamily\bfseries Messidor} & \multicolumn{2}{c||}{\footnotesize\sffamily\bfseries RLDR} & \multicolumn{2}{c}{\footnotesize\sffamily\bfseries Average} \\
& \scriptsize\sffamily\bfseries ACC $\boldsymbol{\uparrow}$ & \scriptsize\sffamily\bfseries F1 $\boldsymbol{\uparrow}$ & \scriptsize\sffamily\bfseries ACC $\boldsymbol{\uparrow}$ & \scriptsize\sffamily\bfseries F1 $\boldsymbol{\uparrow}$ & \scriptsize\sffamily\bfseries ACC $\boldsymbol{\uparrow}$ & \scriptsize\sffamily\bfseries F1 $\boldsymbol{\uparrow}$ & \scriptsize\sffamily\bfseries ACC $\boldsymbol{\uparrow}$ & \scriptsize\sffamily\bfseries F1 $\boldsymbol{\uparrow}$ & \scriptsize\sffamily\bfseries ACC $\boldsymbol{\uparrow}$ & \scriptsize\sffamily\bfseries F1 $\boldsymbol{\uparrow}$ & \scriptsize\sffamily\bfseries ACC $\boldsymbol{\uparrow}$ & \scriptsize\sffamily\bfseries F1 $\boldsymbol{\uparrow}$ & \scriptsize\sffamily\bfseries ACC $\boldsymbol{\uparrow}$ & \scriptsize\sffamily\bfseries F1 $\boldsymbol{\uparrow}$ \\
\midrule[0.8pt]
\multicolumn{3}{l}{\scriptsize\sffamily\bfseries\textit{Domain Generalization Methods}} \\
\midrule[0.8pt]
Mixup  \cite{zhang2018mixup} & 49.4 & 30.2 & 49.7 & 33.3 & 5.8 & 7.4 & 64.0 & 32.6 & 63.0 & 32.6 & 27.7 & 27.0 & 43.3 & 27.2 \\
\rowcolor{gray!5} MixStyle  \cite{zhou2021domain} & 48.8 & 25.0 & 32.0 & 14.6 & 7.0 & 7.9 & 53.5 & 19.4 & 57.6 & 16.8 & 18.3 & 6.4 & 36.2 & 15.0 \\
DDAIG  \cite{zhou2020domain} & 48.7 & 31.6 & 38.5 & 29.7 & 5.0 & 5.5 & 60.2 & \textbf{\cellcolor{yellow!30}33.4} & \textbf{\cellcolor{red!30}69.1} & 35.6 & 25.4 & 23.5 & 41.2 & 26.6 \\
\rowcolor{gray!5} ATS  \cite{sun2020test} & 51.7 & 32.4 & 52.4 & 33.5 & 5.3 & 5.7 & 66.6 & 30.6 & 64.8 & 32.4 & 24.2 & 23.9 & 44.2 & 26.4 \\
Fishr  \cite{rame2022fishr} & \textbf{\cellcolor{yellow!30}61.7} & 31.0 & \textbf{\cellcolor{red!30}61.0} & 30.1 & 6.0 & 7.2 & 48.0 & 30.6 & 52.0 & 33.8 & 19.3 & 21.3 & 41.3 & 25.7 \\
\rowcolor{gray!5} MDLT  \cite{yang2022multi} & 53.3 & 32.4 & 50.2 & 33.7 & 7.1 & 7.8 & 61.7 & 30.6 & 58.9 & 34.1 & 29.0 & 30.0 & 43.4 & 28.1 \\
\midrule[0.8pt]
\multicolumn{3}{l}{\scriptsize\sffamily\bfseries\textit{SOTA Methods for DR Grading}} \\
\midrule[0.8pt]
\rowcolor{gray!5} GREEN  \cite{li2019green} & 52.6 & 33.3 & 44.6 & 31.1 & 5.7 & 6.9 & 60.7 & 33.0 & 54.5 & 33.1 & 31.9 & 27.8 & 41.7 & 27.5 \\
CABNet  \cite{zhou2021cab} & 52.2 & 30.8 & \textbf{\cellcolor{green!30}55.4} & 32.0 & 6.1 & 7.5 & 62.7 & 31.7 & 63.8 & 35.3 & 23.0 & 25.4 & 43.9 & 27.1 \\
\rowcolor{gray!5} MIL-ViT  \cite{bi2023mil} & \textbf{\cellcolor{green!30}61.8} & \textbf{\cellcolor{yellow!30}36.8} & 38.2 & \textbf{\cellcolor{green!30}36.3} & \textbf{\cellcolor{green!30}8.7} & \textbf{\cellcolor{green!30}9.3} & \textbf{\cellcolor{green!30}68.6} & 31.1 & \textbf{\cellcolor{green!30}67.7} & \textbf{\cellcolor{green!30}40.7} & 28.1 & 34.5 & \textbf{\cellcolor{yellow!30}45.5} & 31.5 \\
DRGen  \cite{atwany2022drgen} & 60.7 & 35.7 & 39.4 & 31.6 & 6.8 & 8.4 & 67.7 & 30.6 & 64.5 & 37.4 & 19.0 & 21.2 & 43.0 & 27.5 \\
\rowcolor{gray!5} GDRNet  \cite{che2023towards} & 52.8 & 35.2 & 40.0 & \textbf{\cellcolor{yellow!30}35.0} & 7.5 & \textbf{\cellcolor{yellow!30}9.2} & \textbf{\cellcolor{red!30}70.0} & \textbf{\cellcolor{green!30}35.1} & \textbf{\cellcolor{yellow!30}65.7} & \textbf{\cellcolor{yellow!30}40.5} & \textbf{\cellcolor{green!30}44.3} & \textbf{\cellcolor{green!30}37.9} & \textbf{\cellcolor{green!30}46.7} & \textbf{\cellcolor{yellow!30}32.2} \\
CLIP-DR  \cite{yu2024clip} & 46.3 & 31.8 & 45.8 & 32.6 & 7.8 & \textbf{\cellcolor{green!30}9.3} & 41.9 & 28.3 & 47.3 & 32.5 & \textbf{\cellcolor{yellow!30}41.0} & \textbf{\cellcolor{yellow!30}35.2} & 38.4 & 28.3 \\
\midrule[0.8pt]
\multicolumn{3}{l}{\scriptsize\sffamily\bfseries\textit{Prototype Learning Methods}} \\
\midrule[0.8pt]
ViT-B/16  \cite{dosovitskiy2021image} & 38.2 & 24.6 & 30.1 & 20.5 & 5.4 & 5.6 & 42.3 & 21.8 & 48.5 & 25.7 & 18.6 & 15.2 & 30.5 & 18.9 \\
\rowcolor{gray!5} ProtoNet  \cite{snell2017prototypical} & 52.3 & 33.7 & 39.8 & 27.4 & 7.8 & 7.2 & 54.1 & 28.9 & 58.2 & 32.5 & 28.9 & 24.6 & 40.2 & 25.7 \\
SemFew  \cite{zhang2024simple} & 45.6 & 29.3 & 35.2 & 23.8 & 5.7 & 6.4 & 48.5 & 25.3 & 52.9 & 28.7 & 24.1 & 19.8 & 35.3 & 22.2 \\
\rowcolor{gray!5} LGPN  \cite{liu2025guiding} & 60.4 & \textbf{\cellcolor{green!30}51.2} & 45.7 & 34.3 & \textbf{\cellcolor{yellow!30}8.4} & 6.5 & 54.8 & 32.7 & 55.7 & 39.8 & 35.8 & 33.4 & 43.5 & \textbf{\cellcolor{green!30}33.0 }\\
\midrule[0.8pt]
\rowcolor{gray!20} Ours & \textbf{\cellcolor{red!30}63.1} & \textbf{\cellcolor{red!30}55.3} & \textbf{\cellcolor{yellow!30}52.7} & \textbf{\cellcolor{red!30}41.9} & \textbf{\cellcolor{red!30}11.2} & \textbf{\cellcolor{red!30}9.7} & \textbf{\cellcolor{yellow!30}63.4} & \textbf{\cellcolor{red!30}35.4} & 65.2 & \textbf{\cellcolor{red!30}45.3} & \textbf{\cellcolor{red!30}45.1} & \textbf{\cellcolor{red!30}38.8} & \textbf{\cellcolor{red!30}50.1} & \textbf{\cellcolor{red!30}37.7} \\
\bottomrule[1.5pt]
\end{tabular}
}
\end{table*}

\vspace{0.5mm}\noindent\textbf{Pathological Semantic Injector (PSI).} The first stage integrates diversified description features $\mathbf{E}'$ into the initial prototypes $\mathcal{P}$ to obtain enhanced prototypes $\dot{\mathcal{P}} = \{\dot{\mathcal{P}}_c\}_{c \in \mathcal{C}}$ with richer semantics. We formalize this integration process through an attention-based fusion mechanism. For each severity grade $c \in \mathcal{C}$, the enhancement is computed as:
\begin{equation}
\mathbf{Q_1} = \mathcal{P}_c W_{q1}, \mathbf{K_1} = \mathbf{E}'_c W_{k1}, \mathbf{V_1} = \mathbf{E}'_c W_{v1}.
\end{equation}
\begin{equation}
\mathbf{Q_1} \in \mathbb{R}^{N_s \times d_p}, \mathbf{K_1} \in \mathbb{R}^{N_{\text{div}} \times d_p}, \mathbf{V_1} \in \mathbb{R}^{N_{\text{div}} \times d_v}.
\end{equation}
\begin{equation}
\dot{\mathcal{P}}_c = \mathcal{P}_c + \textit{softmax}\left(\frac{\mathbf{Q_1} \mathbf{K_1}^T}{\sqrt{d_p}}\right)\mathbf{V_1}, \quad c \in \mathcal{C},
\end{equation}
Here, $W_{q1} \in \mathbb{R}^{d_v \times d_p}$, $W_{k1} \in \mathbb{R}^{d_t \times d_p}$ and $W_{v1} \in \mathbb{R}^{d_t \times d_v}$ are learnable parameters of fully connected layers, $N_s$ represents the number of visual tokens, $N_{\text{div}}$ denotes the number of diverse prompts, and $d_p$ represents the projection feature dimension. 

The attention mechanism enables selective and adaptive integration of DR pathological features $\Phi = \{\phi_1, \phi_2, ..., \phi_m\}$ (such as microaneurysms, hard exudates, intraretinal hemorrhages, and other abnormalities) into the visual token representation. The attention weights $A \in \mathbb{R}^{N_s \times N_{\text{div}}}$ quantify the relevance between each visual token and semantic description, enabling precise mapping from macro-semantic descriptions to micro-pathological regions. Consequently, the enhanced prototype $\dot{\mathcal{P}}_c$ can more accurately express the key pathological features of various DR grades.

\vspace{0.5mm}\noindent\textbf{Discriminative Prototype Enhancer (DPE)}. In the second stage, we integrate differentiated description features $\mathbf{E}''$ into enhanced prototypes $\dot{\mathcal{P}}$ to obtain final prototypes $\ddot{\mathcal{P}} = \{\ddot{\mathcal{P}}_c\}_{c \in \mathcal{C}}$ with discriminative information. We formulate this enhancement as:
\begin{equation}
\begin{aligned}
\ddot{\mathcal{P}}_c = \dot{\mathcal{P}}_c + \text{LayerNorm}(\textit{softmax}\left(\frac{\mathbf{Q_2} \mathbf{K_2}^T}{\sqrt{d_p}}\right)\mathbf{V_2}), \quad c \in \mathcal{C},
\end{aligned}
\end{equation}
where the query and key matrices are computed as:
\begin{equation}
\begin{aligned}
\mathbf{Q_2} &= \dot{\mathcal{P}}_c W_{q2} \in \mathbb{R}^{N_s \times d_p}, \\
\mathbf{K_2} &= \mathbf{E}''_c W_{k2} \in \mathbb{R}^{N_{\text{diff}} \times d_p}.
\end{aligned}
\end{equation}
Here, $W_{q2} \in \mathbb{R}^{d_v \times d_p}$ and $W_{k2} \in \mathbb{R}^{d_t \times d_p}$ are learnable projection matrices, and $N_{\text{diff}}$ represents the number of differentiated descriptions per grade pair. The LayerNorm helps ensure the stability of training by normalizing the semantic features of different levels.

To reduce the influence of irrelevant features that do not contribute to grade differentiation, we introduce an adaptive weighting mechanism to calculate the value matrix $\mathbf{V_2}$:
\begin{equation}
\mathbf{V_2} = \sum_{c' \in \mathcal{C} \setminus \{c\}} \alpha_{c,c'} \cdot (\mathbf{E}''_{c,c'} W_{v2}),
\end{equation}
where $W_{v2} \in \mathbb{R}^{d_t \times d_v}$ is a learnable parameter matrix, and $\alpha_{c,c'}$ is an adaptive weight calculated as:
\begin{equation}
\alpha_{c,c'} = \sigma\left(\frac{1}{N_s}\sum_{i=1}^{N_s}\frac{(\dot{\mathcal{P}}_c[0] W_p) \cdot (\mathbf{E}'_{c,c'})^T}{\sqrt{d_t}}\right).
\end{equation}
Here, $\sigma(x) = \frac{1}{1+e^{-x}}$ is the Sigmoid function that maps values to the range $(0,1)$, $W_p \in \mathbb{R}^{d_v \times d_t}$ implements linear mapping from vision to text space, and $\dot{\mathcal{P}}_c[0] \in \mathbb{R}^{1 \times d_v}$ represents the global feature (using the [CLS] token) of the enhanced prototype. This LVLM-driven adaptive weighting mechanism quantifies the importance of differentiated descriptions between each pair of grades. Ultimately, the DPE module enhances the discriminative ability of prototypes by integrating differentiated information, establishing clearer decision boundaries between adjacent DR severity grades.

\subsection{Optimization and Inference}
Given a query image feature $\mathbf{X}_q \in \mathbb{R}^{N_s \times d_v}$ and the semantically enhanced prototype set $\ddot{\mathcal{P}} = \{\ddot{\mathcal{P}}_c\}, {c \in \mathcal{C}}$, the classification process is achieved by computing the similarity between the features and each prototype. Formally, the classification loss is defined as:
\begin{equation}
\mathcal{L}_{cls} = -\log \frac{\exp(\langle \mathbf{X}_q, \ddot{\mathcal{P}}_{y_q} \rangle / \tau)}{\sum_{c \in \mathcal{C}} \exp(\langle \mathbf{X}_q, \ddot{\mathcal{P}}_c \rangle / \tau)},
\end{equation}
where $\langle \cdot, \cdot \rangle$ represents the cosine similarity operation, $y_q \in \mathcal{C}$ is the true label, and $\tau$ is the temperature parameter. During the model inference phase, given a test image feature $\mathbf{X}_{test}$, its classification prediction is computed as:
\begin{equation}\hat{y} = \arg\max_{c \in \mathcal{C}} \langle \mathbf{X}_{test}, \ddot{\mathcal{P}}_c \rangle.
\end{equation}
This framework achieves progressive learning from vision to semantics to discriminative features, effectively bridging the gap between visual representations and medical semantic knowledge.

\section{Experiments}
\subsection{Experimental Setup}
\vspace{0.6mm}
\textbf{Datasets.}
To comprehensively evaluate the effectiveness of the proposed method in cross-domain diabetic retinopathy (DR) grading tasks, we conducted experiments on eight public datasets that span different regions, imaging devices, and annotation standards. These include EyePACS \cite{diabetic-retinopathy-detection}, MESSIDOR \cite{decenciere2014feedback}, IDRiD \cite{porwal2020idrid}, APTOS\cite{aptos2019-blindness-detection}, DeepDR \cite{zeng2019deepdr}, FGADR \cite{zhou2020benchmark}, RLDR \cite{zhou2020benchmark}, and DDR \cite{li2019diagnostic}. These datasets provide an extensive foundation for evaluation.

\vspace{0.6mm}
\noindent\textbf{Cross-domain Evaluation Setup.}
Following prior works \cite{xie2023towards, yu2024clip}, we evaluated our method using two cross-domain generalization scenarios and the aforementioned public datasets. The first scenario, Extreme Single Domain Generalization (ESDG), used a single training dataset and evaluated the model on DDR \cite{li2019diagnostic} and EyePACS \cite{diabetic-retinopathy-detection} to assess its generalization ability. The second scenario, the classic leave-one-out cross-domain test (DG), used one dataset of the six datasets in GDRBench \cite{yu2024clip} as the target domain and the others as source domains to evaluate the model's transfer capability.
In both scenarios, we maintained consistent preprocessing protocols and metrics to ensure fair comparisons with baseline methods.


\vspace{0.6mm}
\noindent\textbf{Implementation Details.}
We employed a series of preprocessing and training strategies to optimize model performance. All fundus images were resized to 224$\times$224 pixels. The model architecture used ViT-B/16 pretrained on SSIT \cite{huang2024ssit} as the feature extractor, with the token count $N_s$ set to 197 and the dimension $d_p$ to 384. The hidden size of the text encoder was 768. We set $\alpha = 5$, $N_{div} = 11$, $N_{diff} = 4$, batch size to 128, and $\tau = 1$. For semantic representation, GPT-4 \cite{achiam2023gpt} generated descriptions and differential text for DR, while CLIP-DR (ViT-B/16 with 768-dimensional encoding) \cite{yu2024clip} acted as the LVLM. Training was performed with a batch size of 32 using the Adam optimizer and an initial learning rate of 1e-4 for 50 epochs, saving the best model based on validation loss. All experiments were conducted on an NVIDIA GeForce RTX3090 GPU (24GB VRAM).

\noindent\textbf{Comparative Methods.}
We evaluated the effectiveness of HAPM by comparing it with three categories of techniques. Domain generalization methods:  Mixup  \cite{zhang2018mixup}, MixStyle \cite{zhou2021domain}, DDAIG \cite{zhou2020domain}, ATS \cite{sun2020test}, Fishr  \cite{rame2022fishr}, MDLT \cite{yang2022multi}; State-of-the-art methods for DR grading: GREEN \cite{li2019green}, CABNet \cite{zhou2021cab}, MIL-ViT \cite{bi2023mil}, DRGen \cite{atwany2022drgen}, and GDRNet \cite{che2023towards}, as well as the CLIP-based cross-domain DR grading method CLIP-DR \cite{yu2024clip}; finally, prototype learning methods: we compared the SSL pre-trained ViT-B/16 architecture \cite{huang2024ssit}, selected the prototype network ProtoNet \cite{snell2017prototypical}, and prototype evolution techniques, including SemFew \cite{zhang2024simple}, which uses semantic evolution to automatically generate high-quality semantic information, and LGPN  \cite{liu2025guiding}, which uses label semantics to guide the prototype network. Model performance was evaluated using \textit{accuracy} and \textit{F1} score.


\begin{figure}[!t]
    \centering
    \includegraphics[width=0.94\linewidth]{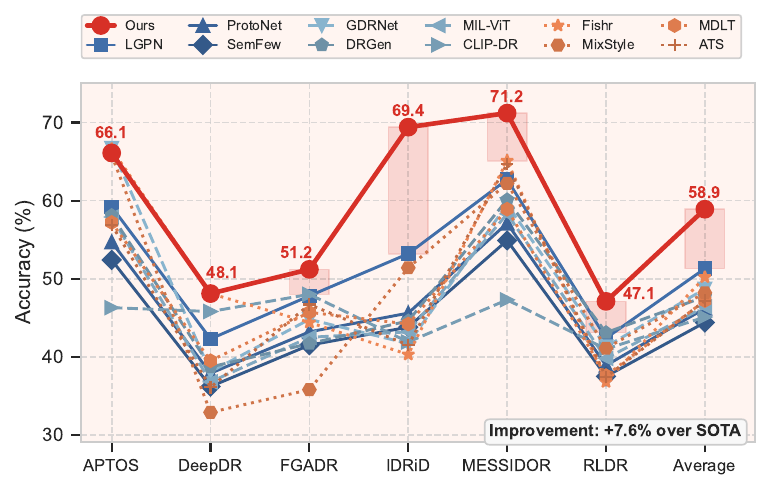}
    \vspace{-2.5mm}
    \caption{The DG performance comparison on six benchmark datasets and average levels. The red areas indicate our method’s performance gain over others on each dataset.}
    \vspace{-1mm}
    \label{fig:dg_performance}
\end{figure}

\subsection{Main Results}
In Table \ref{tab:results} and Figure \ref{fig:dg_performance}, we present the experimental results of our HAPM method across two evaluation settings.

For Extreme Single Domain Generalization, HAPM consistently outperforms existing methods across six datasets, achieving an average accuracy of 50.1\% and an F1 score of 37.7\%, surpassing the second-best methods GDRNet and MIL-ViT. For Leave-one-out Domain Generalization, as shown in Figure \ref{fig:dg_performance}, HAPM outperforms existing methods on all datasets, with an average F1 score of 58.9\% (7.6\% improvement over the second-best method). While methods like Fishr and MixStyle show competitive results, DR-specific methods such as GDRNet and CLIP-DR perform well on specific datasets but struggle with style variations. The prototype-based method LGPN shows good generalization through label guidance. Although HAPM may be slightly inferior on individual metrics, it relies solely on prototypes without full fine-tuning, demonstrating significant overall advantages and validating our prototype modulation strategy for addressing visual-semantic evolution in DR grading, offering a robust solution for clinical practice.

\begin{table}[t]
\centering
\renewcommand{\arraystretch}{0.9}
\setlength{\tabcolsep}{3pt}
\small
\caption{Ablation Study of HAPM Framework Components for Cross-Domain DR Grading on APTOS Dataset.}
\label{ablation1}
\resizebox{0.47\textwidth}{!}{
\begin{tabular}{p{3cm}||c||c||c||c}
\toprule[1.5pt]
 {\small\sffamily\bfseries Method} & \multicolumn{1}{c||}{\small\sffamily\bfseries APTOS} & \multicolumn{1}{c||}{\small\sffamily\bfseries DeepDR} & \multicolumn{1}{c||}{\small\sffamily\bfseries FGADR} & \multicolumn{1}{c}{\small\sffamily\bfseries Average} \\
\midrule[0.8pt]
Base Model & 24.6 & 20.5 & 5.6 & 16.9 \\
\rowcolor{gray!5} \textbf{+}Anchor Prototype & 33.5 & 26.3 & 6.8 & 22.2 \\
\textbf{+}Basic Prompt & 41.2 & 30.9 & 7.5 & 26.5 \\
\rowcolor{gray!5} \textbf{+}PSI Module & 47.8 & 35.2 & 8.6 & 30.5 \\
\textbf{+}DPE \textit{w/o} AdaptWeight & 50.4 & 37.8 & 8.9 & 32.4 \\
\rowcolor{gray!5} \textbf{+}DPE \textit{w/} AdaptWeight & 52.1 & 39.4 & 9.2 & 33.6 \\
\midrule[1.2pt]
 \rowcolor{gray!20}\textbf{Complete HAPM} & \textbf{55.3} & \textbf{41.9} & \textbf{9.7} & \textbf{35.6} \\
\bottomrule[1.5pt]
\end{tabular}
}
\end{table}

\begin{table}[t]
\centering
\caption{Analysis of Different Prompt Sources on Cross-Domain DR Grading Performance.}
\label{ablation2}
\renewcommand{\arraystretch}{0.9}
\setlength{\tabcolsep}{4pt}
\small
\resizebox{0.47\textwidth}{!}{
\begin{tabular}{l||c||c||c||c||c}
\toprule[1.5pt]
\textbf{Prompt Source} & \textbf{\#Prompts} & \textbf{APTOS} & \textbf{DeepDR} & \textbf{FGADR} & \textbf{Average} \\
\midrule[0.8pt]
LVLM Prompts & 5 (Manual) & 48.8 & 37.4 & 8.6 & 31.6 \\
\midrule[0.5pt]
\multirow{3}{*}{GPT-4 Generated} & 25 (Auto) & 50.3 & 38.2 & 8.8 & 32.4 \\
& 50 (Auto) & 52.0 & 39.7 & 9.1 & 33.6 \\
& 75 (Auto) & 51.6 & 39.3 & 9.0 & 33.3 \\
\midrule[0.5pt]
\multirow{2}{*}{Hybrid Prompts} & 30 (5M+25A) & 52.7 & 40.1 & 9.2 & 34.0 \\
& 55 (5M+50A) & 53.8 & 40.8 & 9.4 & 34.7 \\
\midrule[0.8pt]
\rowcolor{gray!20} Hybrid+Diff Desc & 55+20 & \textbf{55.3} & \textbf{41.9} & \textbf{9.7} & \textbf{35.6} \\
\bottomrule[1.5pt]
\multicolumn{6}{l}{\footnotesize Hybrid Prompts: LVLM Prompts + GPT-4 Generated Prompts}\\
\end{tabular}
}
\end{table}

\subsection{Ablation Study}

\vspace{0.5mm}\noindent \textbf{Effect of Each Component.} 
The results in Table \ref{ablation1} validate the effectiveness of each component in our HAPM framework. The variance spectrum-driven anchor prototype selection significantly enhanced the model's ability to capture domain-invariant pathological features, while the basic prompt adaptation mechanism further strengthened semantic integration. The PSI module effectively improved domain generalization capability by injecting disease-specific semantic information into visual prototypes. The DPE module with its adaptive weighting mechanism precisely modulated the discriminative boundaries between adjacent DR levels. The complete HAPM framework achieved a 18.7\% improvement compared to the baseline model, demonstrating the effectiveness of our proposed hierarchical prototype progressive modulation strategy.


\vspace{0.5mm}\noindent \textbf{Effect of Prompt Sources.} Table \ref{ablation2} shows that hybrid prompting strategies (combining manually designed and LLM-generated prompts) significantly outperform single-source prompts. The best results were achieved by combining hybrid prompts with differential descriptions, which confirms that our multi-stage prompting strategy effectively captures subtle features and enhances the model's ability to recognize clinical boundary cases.

\vspace{0.5mm}\noindent \textbf{Effect of Anchor Image Selection Strategy and Quantity.} In Table \ref{ablation4}, the variance-based anchor selection strategy achieved optimal performance when selecting 5 anchors per class, outperforming both random selection and class centroid selection methods. Even with just 1 anchor, the variance-based method still showed strong performance, while increasing anchors to 10 did not improve performance and caused a slight decrease due to potential noise introduction. The results highlight the importance of anchor quality over quantity in cross-domain generalization. Based on these findings, we finally chose 5 anchors as a compromise for experiments.


\begin{table}[t]
\centering
\caption{Impact of Different Anchor Selection Strategies on Cross-Domain DR Grading Performance.}
\label{ablation4}
\renewcommand{\arraystretch}{0.9}
\LARGE 
\resizebox{0.48\textwidth}{!}{
\begin{tabular}{p{3.8cm}||c||cccc}
\toprule[2pt]
 \multirow{2}{*}{\textbf{Strategy}}& \multicolumn{1}{c||}{\multirow{2}{*}{\textbf{\#Anchors}}} &\multicolumn{4}{c}{\textbf{Cross-Domain F1 (\%)}} \\
 & & \textbf{APTOS} & \textbf{DeepDR} & \textbf{FGADR} & \textbf{Average} \\
\midrule[1pt]
\multirow{3}{*}{Random} & 1 & 49.8 & 39.8 & 11.0 & 33.5 \\
& 5 & 50.6 & 40.2 & 11.4 & 34.1 \\
& 10 & 51.2 & 39.9 & 11.3 & 34.1 \\
\midrule[0.5pt]
\multirow{3}{*}{Class Centroid} & 1 & 50.7 & 39.5 & 11.2 & 33.8 \\
& 5 & 52.9 & 41.3 & 11.5 & 35.2 \\
& 10 & 52.5 & 41.9 & 11.4 & 35.3 \\
\midrule[0.5pt]
\multirow{3}{*}{Variance-Based (Ours)} & 1 & 53.8 & 41.7 & 9.5 & 35.0 \\
 & 5 & \textbf{55.3} & \textbf{41.9} & \textbf{9.7} & \textbf{35.6} \\
& 10 & 54.1 & 40.8 & 9.4 & 34.8 \\
\bottomrule[2pt]
\multicolumn{6}{l}{\large Variance-Based: Selects anchors with minimal intra-class feature variance}\\
\end{tabular}
}
\end{table}

\begin{figure}[t]
\centering
\includegraphics[width=1\linewidth]{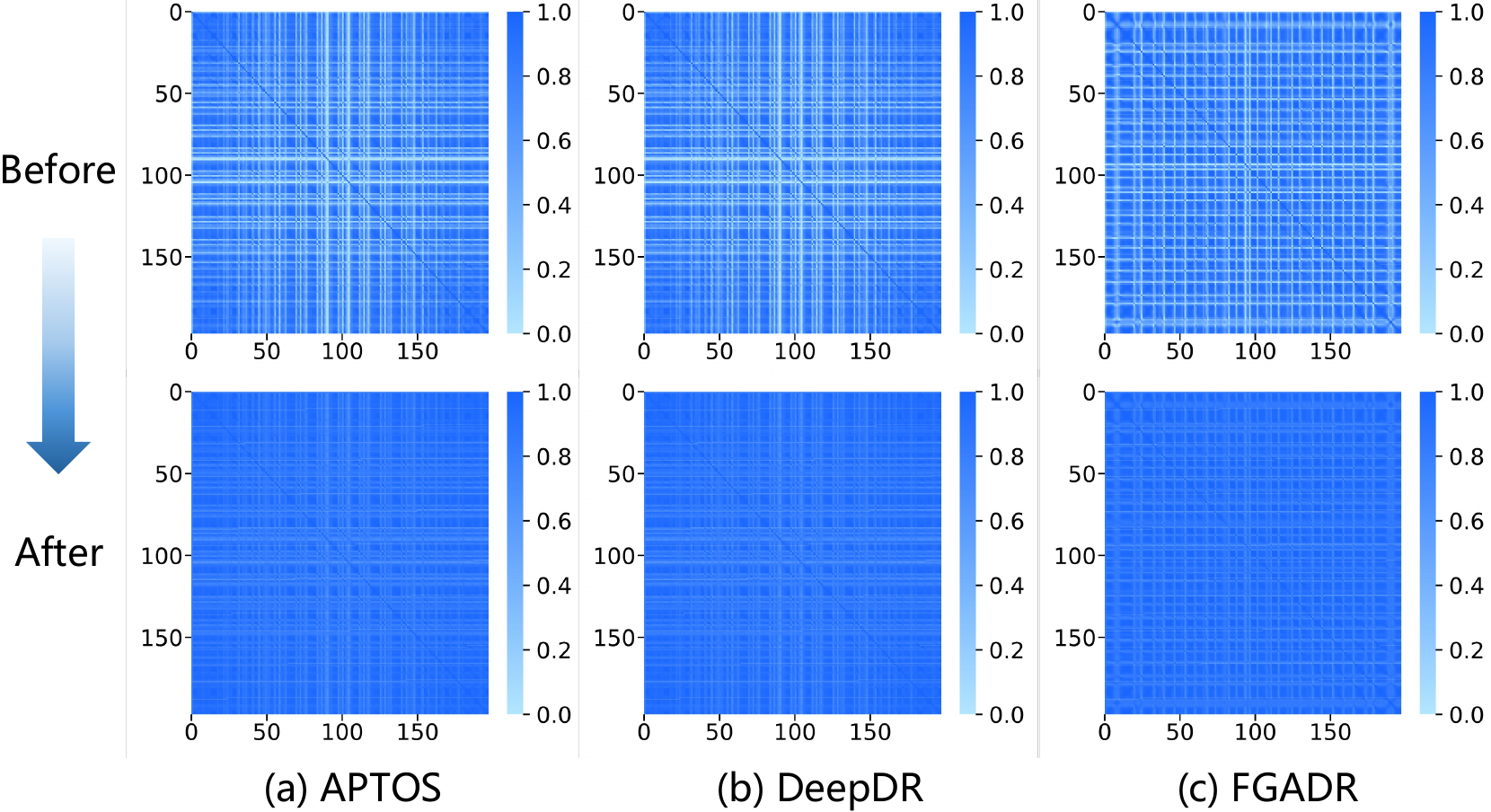}
\caption{Analysis of the prototype token correlation matrices before and after modulation reveals stronger token correlations after our proposed two-stage modulation.}
\label{correlation_matrices}
\end{figure}

\subsection{In-Depth Analysis}
As shown in Figure \ref{correlation_matrices}, the Token Correlation Matrices reveal our prototype modulation strategy's effectiveness across three datasets. Before modulation, token correlations display heterogeneity with scattered high-correlation regions, indicating inconsistent feature relationships. After two-stage modulation, we observe significantly enhanced correlation coherence across all datasets, demonstrating that our PSI and DPE modules successfully integrate clinical knowledge into visual representations. This transformation shows that our framework preserves domain-invariant pathological patterns,  generating more clinically meaningful representations for distinguishing adjacent severity levels in DR grading.

\begin{figure}[t]
\centering
\includegraphics[width=1\linewidth]{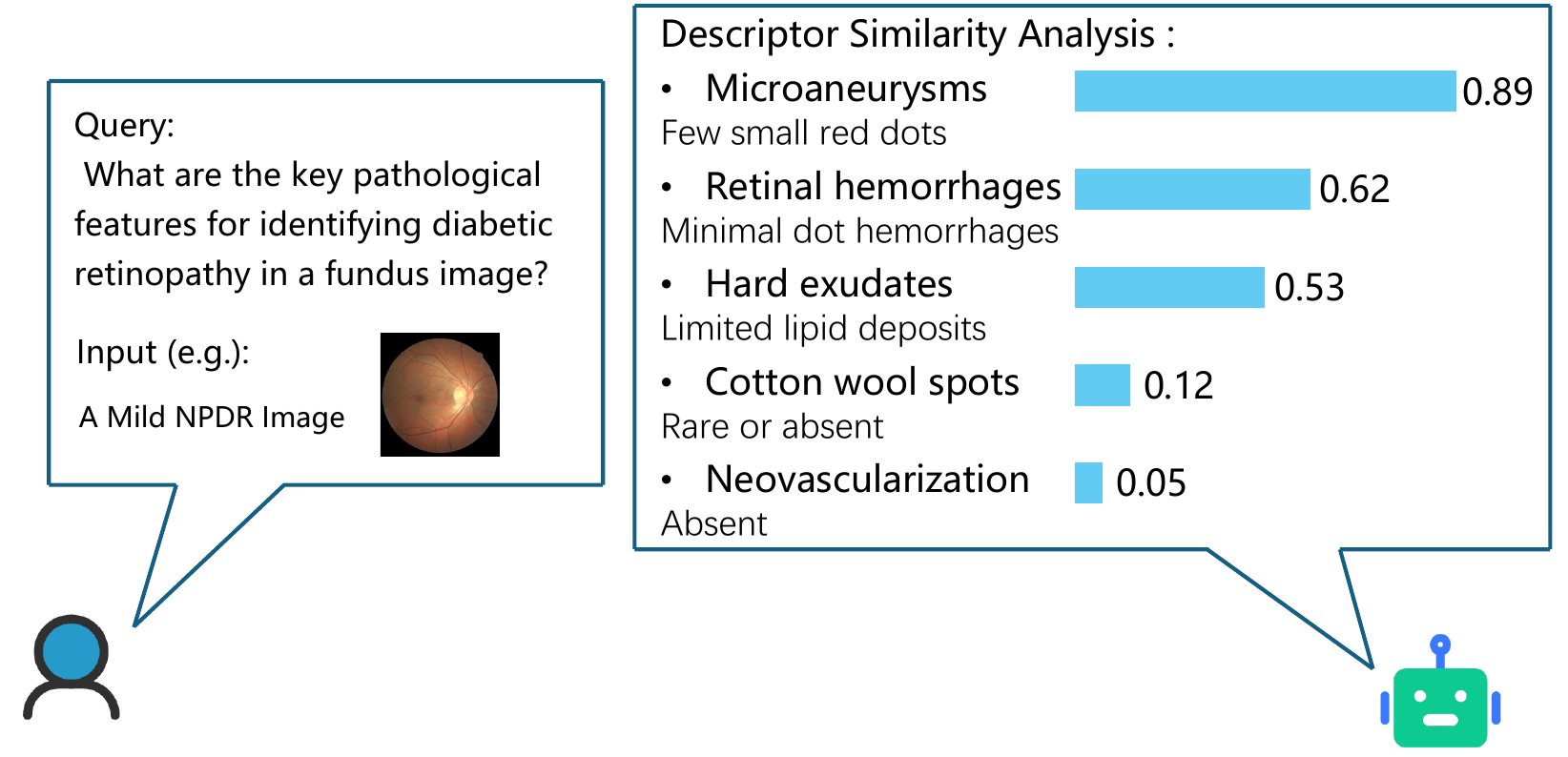}
\caption{LLM-generated descriptor similarity comparison after prototype modulation. The bars show how HAPM enhance the correlation between image and pathological descriptors, particularly strengthening relevant descriptors.}
\label{exp}
\end{figure}

\subsection{Interpretation Study}
We present result for explainable DR grading in Figure \ref{exp}, where bars illustrate descriptor similarity to images classified at different severity levels in the latent space after prototype modulation. We prompted an LLM with: ``\texttt{Q: What are the key pathological features for identifying diabetic retinopathy in a fundus image?}'' to generate five common pathological descriptors across all DR grades: retinal hemorrhages, hard exudates, microaneurysms, cotton wool spots, and neovascularization. After applying our PSI and DPE modules, we observe enhanced alignment between descriptors and image representations, revealing clearer, clinically meaningful patterns: for example, mild NPDR shows stronger correlation with microaneurysms and greater distinction from proliferative indicators. This modulated descriptor relevance demonstrates both improved hierarchical pattern alignment with clinical progression and enhanced inter-class separation, effectively validating our framework's ability to ground classifications in clinically interpretable pathological evidence.

%
%
%

\vspace{1mm}
\section{Conclusion}
\vspace{1mm}
This paper introduces the Hierarchical Anchor Prototype Modulation (HAPM) framework for diabetic retinopathy grading that addresses adjacent levels confusion through pathology-aware evolution. By implementing variance spectrum-driven anchor selection, hierarchical prompt gating, and a two-stage prototype modulation strategy, HAPM effectively drives visual prototype evolution through pathological descriptions and differential characterizations for accurate DR severity classification. Experimental results across multiple datasets demonstrate superior performance in distinguishing borderline cases between adjacent DR grades. Future research could further explore approaches for low-quality images and leverage generative AI  techniques to assist in challenging scenarios.


\newpage
\clearpage
\begin{acks}
This research was partially supported by grants from the National Natural Science Foundation of China (Grants No. 62472157, No.62202\\158, No.62206089), and the Science and Technology Innovation Program of Hunan Province (Grant No. 2023RC3098).
\end{acks}
\bibliographystyle{ACM-Reference-Format}
\bibliography{main}

\end{document}